\documentclass[sigconf]{acmart}

\usepackage{booktabs} 
\usepackage{color}
\usepackage{amsmath}
\usepackage[skip=0.4\baselineskip]{caption}

\setcopyright{rightsretained}

\acmDOI{10.475/123_4}

\acmISBN{123-4567-24-567/08/06}

\acmConference[WOODSTOCK'97]{ACM Woodstock conference}{July 1997}{El
  Paso, Texas USA}
\acmYear{1997}
\copyrightyear{2016}

\acmArticle{4}
\acmPrice{15.00}

\begin{document}
\title{Infant Mortality Prediction using Birth Certificate Data}


\author{Antonia Saravanou}
\affiliation{%
   \institution{Department of Informatics and Telecommunications\\ University of Athens}
}
\email{antoniasar@di.uoa.gr}

\author{Clemens Noelke}
\affiliation{%
   \institution{The Heller School for Social Policy and Management\\ Brandeis University}
}
\email{cnoelke@brandeis.edu}

\author{Nicholas Huntington}
\affiliation{%
   \institution{The Heller School for Social Policy and Management\\ Brandeis University}
}
\email{nhuntington@brandeis.edu }

\author{Dolores Acevedo-Garcia}
\affiliation{%
   \institution{The Heller School for Social Policy and Management\\ Brandeis University}
}
\email{dacevedo@brandeis.edu }

\author{Dimitrios Gunopulos}
\affiliation{%
   \institution{Department of Informatics and Telecommunications\\ University of Athens}
}
\email{dg@di.uoa.gr}

\renewcommand{\shortauthors}{A. Saravanou et al.}

\begin{abstract}
The Infant Mortality Rate (IMR) is the number of infants per 1000 that do not survive until their first birthday. It is an important metric providing information about infant health but it also measures the society's general health status. Despite the high level of prosperity in the U.S.A., the country's IMR is higher than that of many other developed countries. Additionally, the U.S.A. exhibits persistent inequalities in the IMR across different racial and ethnic groups~\cite{cdc}.
In this paper, we study the infant mortality prediction using features extracted from birth certificates. We are interested in training classification models to decide whether an infant will survive or not. We focus on exploring and understanding the importance of features in subsets of the population; we compare models trained for individual races to general models. 
Our evaluation shows that our methodology outperforms standard classification methods used by epidemiology researchers.

\end{abstract}

\maketitle

\section{Introduction}
\label{sec:intro}

An important measurement for the overall health level in a society is the infant mortality rate (IMR). 
The IMR in the U.S.A. is higher than in many other developed countries. More specifically, among~26 highly developed countries, the U.S.A. ranks last with IMR=$6.1$, which is nearly 2.6-times larger than the IMR in Japan or Finland\footnote{\url{https://www.cdc.gov/nchs/data/nvsr/nvsr63/nvsr63_05.pdf}}.

Understanding the problem of infant mortality and identifying the impact of several variables is an important yet complex undertaking that requires the active engagement of several scientific fields. In addition to analyzing the factors that are more important on predicting infant mortality risk, our findings could be help to the policy makers. 
If this information can be used to predict infant mortality beyond the clinic, it could provide vital information for effective targeting of social services to children with an elevated mortality risk, such as nurse home visits.

\begin{figure}[t]
\centering
\includegraphics[width=0.95\linewidth]{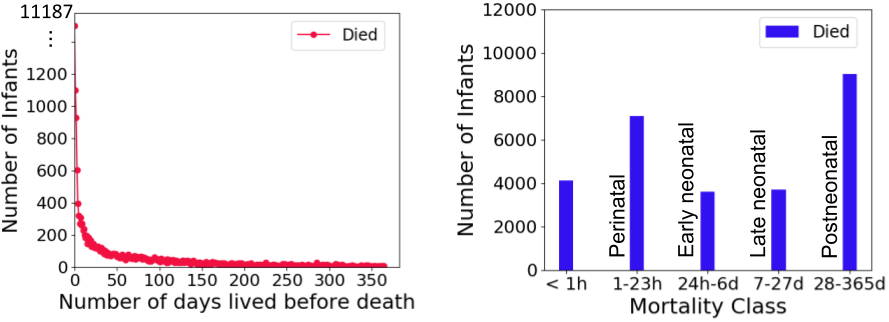}
\caption{(Left) The number of infants who died on the \textit{i}-th day since their birth. (Right) The number of infant deaths grouped by the time of death. }
    \label{fig:intro}
\end{figure}


Existing research has used information that is routinely gathered at birth to determine the relative strength of multiple predictors of infant mortality. \textit{Low birth weight}~\cite{ACEVEDOGARCIA_low_birth_weight_education,wilcox1992birth,almond2005costs,hessol1998risk}, \textit{preterm birth}~\cite{hegyi1998apgar,dolores_hypersegregated_areas} and the \textit{APGAR} score~\cite{casey2001continuing,doyle2003race} have emerged as strong medical risk factors. 
More specifically, apgar score is a measure calculated within the first five minutes of newborn's life and it can quickly summarize her overall health, by evaluating the following five criteria: Appearance, Pulse, Grimace, Activity and Respiration. 
Other studies have established large effects of socio-economic factors~\cite{Acevedo_Garcia2005,dolores_hypersegregated_areas,powers2006race,hessol1998risk}, such as\textit{ maternal race/ethnicity} or \textit{education}. 
However, to our knowledge, there has been no attempt made to predict infant mortality using a combination of the medical measurements together with an extensive number of socioeconomic risk factors. 
Additionally, these existing approaches require constant monitoring of infants which limits their application at large population.

In this work, we are interested to discover how socio-economic factors, e.g., the race or the education of the mother, can improve predicting infant mortality. The majority of infant deaths occur close to the birth date and up to a week after because of complications that appear during the pregnancy or right after the birth (see Figure~\ref{fig:intro}). However, that is not the case for deaths occurring after the infant leaves the hospital. We propose to leverage external information related to the socio-economic status of the infant's family in order to improve the results. 

Our contributions are the following: {(i)} we define the infant mortality problem and e propose models to solve this binary classification task. This task is challenging due to the high imbalance between the two classes (1:145), {(ii)} we investigate the effectiveness of our model in predicting the mortality classes and the causes of death, {(iii)} using external features related to the mother of the infant, we train models for slices of the population (e.g., based on race), {(iv)} we discover the most critical features for our problem, and, {(v)} our evaluation on real-world data shows that the proposed methodology outperforms the existing one.

\section{Data Description}
\label{sec:dataset}

Through our collaboration with the Institute for Child, Youth and Family Policy at the Heller School for Social Policy and Management, Brandeis University we have access to datasets with complete birth certificate information from the most recent consecutive years (2000--2002) in the United States~\cite{clemens2019}. Health professionals at the time of the delivery add information to the birth certificates. State public health agencies collect this information from the medical provider and submit the data to the National Center of Health Statistics who makes the data available for research. Our dataset includes $12$M observations, $83$K of which refer to infants that did not survive until their first birthday. This extremely skewed class distribution (1:145) makes the prediction task very difficult.

Each birth certificate includes 128 variables, with information about the infant and its parents. More specifically, information in a birth certificate can be summarized in: the date and place of birth, information about parents, pregnancy details, medical risk factors, obstetric procedures, complications/characteristics of labor and/or delivery, details of the delivery method, maternal morbidity, abnormal conditions of the newborn, and the sex of the newborn. Furthermore, there are 3 features related to mortality; whether the infant survives or not, the infant's age at death (in days), and the cause of death (when linked with the birth certificate).

\section{Classification Methodology}
\label{sec:method}

We are interested in predicting infant mortality, using birth certificate records. We use a set of observations $X_i$ (the variables from a birth certificate) to predict the outcome $Y_i$ (whether the infant is in the \textsc{Survival} class or not) using classification models, where:
\[ Y_i = 
\begin{cases}
    0, & \text{if infant $i$ belongs to \textsc{Survival} class},\\
    1, & \text{otherwise}.
\end{cases}
\]

We train models with training sets of different distribution in the two classes (\textsc{Survival}--\textsc{NotSurvival}) and we evaluate them using test sets with the real distribution (more details in Section~\ref{sec:training}). We use (i) \textbf{GNB}: Gaussian Naive Bayes~\cite{gnb}, (ii) \textbf{SVM}: one-class SVM~\cite{svm} with RBF kernel, (iii) \textbf{XGB}: Boosted Trees~\cite{xgb}, and (iv) \textbf{NN}: a 3-fully-connected-layers neural network (more details in~\ref{sec:comparison}).

In this paper, we compare our models for different mortality classes and causes of death. Also, we study whether socio-economic conditions (e.g., mother's race) play an important role in this classification task. In Section~\ref{sec:race}, we show the results for individual models for each race. In Section~\ref{sec:feature}, we show our findings for the most important factors in infant mortality prediction.

\section{Experimental Setup}
\label{sec:setup}

\subsection{Training}
\label{sec:training}

Since we work in real-world settings with a high imbalanced class distribution (1:145), it is critical that we build models for these settings. We train our models using birth certificates from 2000-2001, and we test and evaluate them on birth certificates from 2002. To overcome the high imbalance issue, we follow a random sampling approach to create less skewed training sets. In each training dataset we keep all instances from the minority class (\textsc{NotSurvival}) and we sample: (i) same number of \textsc{Survival} class instances, (ii) ten times more \textsc{Survival} class instances, and (iii) all \textsc{Survival} class instances. In other words, we create training sets using all \textsc{NotSurvival} class instances and three different distributions of the \textsc{Survival} class:
\begin{itemize}
    \item The \textbf{1:1} dataset with 55.5K \textsc{NotSurvival} class instances and 55.5K \textsc{Survival} class.
    \item The \textbf{1:10} dataset with 55.5K \textsc{NotSurvival} class instances and 548.4K \textsc{Survival} class.
    \item The \textbf{1:145} dataset with 55.5K \textsc{NotSurvival} class instances and 8M \textsc{Survival} class.
\end{itemize}
Some of the birth certificates in our dataset are not complete or might have missing values. To make sure our models are applicable in real scenarios, we include records with missing values and replace them with the mean value for each feature. To test our models, we always use the true imbalanced dataset (1:145) from 2002.

To build our classification models, we tune the parameters of our models using 5-fold cross-validation to maximize \textit{recall} for the minority class, as we are more interested on measuring the ability of the classifier to find all the infants in high risk. Due to lack of space, we do not show the results from the parameter tuning experiments.

\subsection{Evaluation Metrics}
\label{sec:evaluation}

We choose to evaluate our models using Precision-Recall (for the \textsc{NotSurvival} class) and Area Under the Curve (AUC). We do not use Accuracy as it is not appropriate for such an imbalanced dataset.

\subsection{Comparison Models}
\label{sec:comparison}

In Section~\ref{sec:intro}, we study the most common features used in infant mortality prediction according to the literature, those are: (i) the birth weight of the infant (\textbf{BW}), and (ii) the apgar score (\textbf{AP}). 
We evaluate the performance of the models trained by variations of the features, independently and together. We run four combinations: (i) one using all features (\textsc{\textbf{ALL}}), (ii) one using both the birth weight and the apgar score (\textbf{BWAP}), (iii) one using only the birth weight (\textbf{BW}), and (iv) one using only the apgar score (\textbf{AP}).

In medicine and epidemiology, researchers use Linear and Logistic Regression to predict the infant mortality. We use the following comparison models: (i) \textbf{LAS}: Linear Regression (L1), (ii) \textbf{RID}: Linear Regression (L2), and (iii) \textbf{LOG}: Logistic Regression.

As described in Section~\ref{sec:method}, we use: (i) \textbf{GNB}: Gaussian Naive Bayes, (ii) \textbf{SVM}: One-class SVMs with RBF kernel and $\gamma$ = $1\times10^{-9}$, (iii) \textbf{XGB}: Boosted Trees with maximum depth = $3$, minimum child weight = $3$, subsample rate = $0.7$, learning rate = $0.01$, number of estimators = $1250$, and colsample by tree = $0.8$, and (iv) \textbf{NN}: 3-hidden-layers neural network, where the hidden layers use the rectified linear activation function, the output layer uses the sigmoid activation function, and the model is optimized using the binary cross entropy loss function and the Adam version of gradient descent.




\section{Results and Discussion}
\label{sec:results}


\subsection{Binary Classification Task}
\label{sec:binary}

\begin{table}[t]
\centering
\small
\caption{Best performing models for each train set distribution. The same test set (1:145) is used to evaluate all models.}
\begin{tabular}{cccccl}
\toprule
\textbf{Train Set} & \textbf{Features} & \textbf{Method} & \textbf{Precision} & \textbf{Recall} & \textbf{AUC}\\
\midrule
\textsc{1:1} & \textsc{ALL} & $\textsc{LOG}$ &  0.05 &  0.75 & 0.83  \\
\textsc{\textbf{1:10}} & \textsc{ALL} & \textbf{$\textsc{XGB}$}  &  0.06 & \textbf{0.77} & \textbf{0.85}\\
\textsc{1:145} & \textsc{ALL} & {$\textsc{GNB}$} & \textbf{0.07} &  0.67  & 0.80 \\
\bottomrule
\end{tabular}

\label{tbl:all_train}
\end{table}

\begin{table}[t]
\centering
\small
\caption{Evaluation for the classification models using the 1:10 training dataset. Precision and recall values refer only to the minority class. The test set contains 4M birth certificates, in which 27.5K belong to the \textsc{Notsurvival} class.}
\begin{tabular}{clccccccc}
\toprule
\textbf{Features} & \textbf{Metric} & \textbf{{$\textsc{LAS}$}} & \textbf{{$\textsc{RID}$}} & \textbf{{$\textsc{LOG}$}} & \textbf{{$\textsc{GNB}$}} & \textbf{{$\textsc{SVM}$}} & \textbf{{$\textsc{XGB}$}} & \textbf{{$\textsc{NN}$}}\\
\midrule
 & {Prec.} & {0.0} & {0.0} & {0.25} & {0.14} & {0.25} & {0.06} & {0.43}\\
\textsc{BW} & {Recall} & {0.0} & {0.0} & {0.54} & {0.59} & {0.54} & {0.67} & {0.48}\\
 & {AUC} & {0.78} & {0.80} & {0.76} & {0.78} & {0.76} & {0.80} & {0.74}\\

\midrule
 & {Prec.} & {0.0} & {0.0} & {0.0} & {0.0} & {0.0} & {0.05} & {0.25}\\
\textsc{AP} & {Recall} & {0.0} & {0.0} & {0.0} & {0.0} & {0.0} & {0.59} & {0.43}\\
 & {AUC} & {0.5} & {0.5} & {0.5} & {0.5} & {0.5} & {0.76} & {0.71}\\

\midrule

 & {Prec.} & {0.0} & {0.0} & {0.31} & {0.11} & {0.88} & {0.06} & {0.32}\\
\textsc{BWAP} & {Recall} & {0.0} & {0.0} & {0.52} & {0.62} & {0.06} & {0.70} & {0.55}\\
 & {AUC} & {0.80} & {0.80} & {0.76} & {0.79} & {0.53} & {0.81} & {0.77}\\

\midrule

 & {Prec.} & \textbf{0.89} & {0.64} & {0.28} & {0.07} & {0.24} & {0.06} & {0.28}\\
\textsc{All} & {Recall} & {0.15} & {0.32} & {0.56} & {0.68} & {0.54} & \textbf{0.77} & {0.54}\\
 & {AUC} & {0.80} & {0.83} & {0.77} & {0.81} & {0.77} & \textbf{0.85} & {0.76}\\

\bottomrule
\end{tabular}

\label{tbl:bin_res}
\end{table}

For this task, we run the proposed models and the baselines from the literature, with different features combinations (\textsc{All, BWAP, BW, AP}) and training sets (1:1, 1:10, 1:145). We evaluate the models' performance using the same test set that contains the real imbalance (1:145). We show the best performing models for each train set in Table~\ref{tbl:all_train} using \textsc{All} features in the training. Boosted trees with the 1:10 training set achieve the best recall (0.77) and AUC (0.85) scores.

We continue our experimentation using the 1:10 training set and variations of features and models. We show the results in Table~\ref{tbl:bin_res}. In each row (separated by the vertical line) of the table we have the results for all models using the same variation of features. We observe that the best performing setup is using the 1:10 training set, that contains \textsc{All} features from the birth certificate, to train the Boosted trees model with recall = $0.77$ on the minority class and AUC = $0.85$. In such a task, we aim for high recall, as our goal is not to miss any infant in high risk of mortality. Boosted trees outperform the comparison baselines in recall, meaning that the baselines miss a lot of the infant deaths. However, baseline models achieve better precision scores because they are more conservative in predicting in the \textsc{NotSurvival} class. Overall, these results prove that utilizing all features with our parameters setup and Boosted trees outperform all other models-features settings.

\subsection{Discussion}

In this section, we study the predicting performance of our models in different scenarios; more specifically, we analyze the performance for each of the following categories: (i) mortality classes, (ii) causes of death, (iii) races, and (iv) feature importance.

\subsubsection{Evaluation for each mortality class}
\label{sec:mort_class}

\begin{table}[t]
\centering
\small
\caption{Results for each mortality class.}

\begin{tabular}{lcl}
\toprule
\textbf{Mortality class} & \textbf{Accuracy} \\
\midrule

\textbf{< 1 hour} &         \textbf{0.988}  \\
1 - 23 hours &     0.979  \\
1 - 6 days &       0.894  \\
7 - 27 days &      0.767  \\
28 days - 1 year & 0.484  \\	
\bottomrule

\end{tabular}
\label{tbl:ager5}
\end{table}

We are interested to evaluate how the models predict infant deaths from each of the following infant mortality categories: (i) survived less than an hour, (ii) survived between an hour and 23 hours, (iii) survived more than a day and less than a week, (iv) survived more than a week and less than a month, and (v) survived between a month and a year. 

For these series of experiments we used all of our models, but due to space limitation we only report the best performing classifier (Boosted Trees). Here, we analyze how many infants deaths which occur in each mortality category are correctly predicted by our model. We expect that the model achieves better results when the infant death is close to the birth date because all information in birth certificates is up to that date. Our intuition is verified by the results in Table~\ref{tbl:ager5}. Infant deaths that occur after the first month of the newborn are hard to be predicted as the birth certificate records do not continue to update with the infant's health. To improve these results it is important to count on the general family and socio-economic factors for the infant mortality prediction.

\subsubsection{Evaluation for each cause of death}
\label{sec:cause_death}

\begin{table}[t]
\centering
\small
\caption{Results for each cause of death.}

\begin{tabular}{lc}
\toprule
\textbf{Cause of death} & \textbf{Accuracy}\\
\midrule

\textbf{Gestation/Fetal Malnutrition} &          \textbf{0.995}  \\  
SIDS, NEC, External Causes &            0.304  \\  
Congenital/Chromosomal Abnormalities & 0.806  \\  
\textbf{Maternal Factors/Complications} &        \textbf{0.987}  \\  
Diseases/Disorders &                    0.613  \\  
Other &                                 0.937  \\  	
\bottomrule 

\end{tabular}

\label{tbl:cod6}
\end{table}

Moreover, we analyze the predicting accuracy for the infant deaths that are linked to the cause of death, in order to understand if our models perform better in predicting deaths by specific causes. Again, we only report the results from the best performing model (Boosted Trees). We observe that \textsc{XGB} performs well when the death is caused by: \textit{Gestation/Fetal Malnutrition} and \textit{Maternal Factors/Complications}. Those are complications happening during the pregnancy period and the birth and might cause infant death in less than a week. Our model is less accurate when predicting mortality for infants that died because of \textit{Sudden Infant Death Syndrome} (SIDS), \textit{Necrotizing enterocolitis} (NEC), \textit{External Causes or Diseases/Disorders}, which is something expected due to the input information from the birth certificates. 
Again these results show that it is crucial to build models using the socio-economic factors that can be extracted from birth certificates.

\subsubsection{Evaluation for each individual race}
\label{sec:race}

\begin{table}[t]
\centering
\small
\caption{AUC results for the 3 most populated races in the dataset. The column-names refer to the balance distribution in the training set. Testing is done using the real imbalance.}
\begin{tabular}{lcccccl}
\toprule
\textbf{} & \textbf{1:1} & \textbf{1:10} & \textbf{1:145} \\
\midrule
\textbf{White} &  0.84	 & 0.84 &	0.82   \\ 
\textbf{Black} &  0.83 & 0.83 & 0.82   \\  
\textbf{American Indian} & 0.72 & 0.75 & 0.75  \\
\bottomrule
\end{tabular}

\label{tbl:race}
\end{table}

In this work, we are focusing on the impact of the external factors that can improve the predicting performance of infant mortality prediction models. The available features included in the birth certificate that are related to the family of the infant and the socio-economic conditions are limited. 
One of those features, that has been studied in the literature as well~\cite{dolores_hypersegregated_areas,hessol1998risk,hummer1999race,abrevaya2002effects}, is the race of the mother which is also considered as the race of the infant. 
Here, we focus on the mother's race, as paternity is not always acknowledged and might be missing. 
The races found in the birth certificates are: (1) White, (2) Black, (3) American Indian, (4) Chinese, (5) Japanese, (6) Hawaiian, (7) Filipino, (8) Asian Korean, (9) Samoan, (10) Vietnamese, (11) Guamanian, and (12) Other. 
Moving towards our goal, to discover whether a race can improve the prediction of infant mortality, we build models for each individual race. 
We group the birth certificates per race, and we use those from 2000-01 (in the distributions 1:1, 1:10, 1:145) for the training of the models. 
We evaluate the performance of the models using the corresponding test set from 2002 (with birth certificates of the same race). 
In Table~\ref{tbl:race}, we report the AUC results of the top-3 most populated races in the dataset (White, Black, American Indian) using the best performing model (\textsc{XGB}). 
The results show that training different classifiers for each individual race improves the predicting accuracy of the model.

\subsubsection{Findings for feature importance}
\label{sec:feature}

\begin{table}[t]
\centering
\small
\caption{Top-20 features.}

\begin{tabular}{clcl}
\toprule
1 & Birth weight (in grams) & 11 & Detail live birth order\\
2 & Apgar score & 12 & Other abnormal conditions\\
3 & State of occurrences & 13 & Other congenital anomalies\\
4 & Birth weight (cat.) & \textbf{14} & \textbf{Mother's place of birth}\\
\textbf{5} & \textbf{Mother's state of residence} & \textbf{15} & \textbf{Number of cigarettes/day}\\
6 & Father's age & \textbf{16} & \textbf{Number of prenatal visits} \\
\textbf{7} & \textbf{Mother's education (desc.)}  & \textbf{17} & \textbf{Mother's age} \\
8 & Heart malformations & \textbf{18} & \textbf{Mother's race} \\
9 & Assisted ventilation (>30')  & \textbf{19} & \textbf{Mother's marital status}\\
10 & Gestation (in weeks) & \textbf{20} & \textbf{Mother's education (cat.)}\\
\bottomrule
\end{tabular}
\label{tbl:feat_bin}
\end{table}

In this section, we discuss our findings on the feature importance. In general, the importance of the feature indicates how informative and useful the feature was for building the model. We experiment on the importance of the features for the binary classifier and the individual classifiers for each race, but due to lack of space we report our findings only on the binary classifier. 
In Boosted Trees, the importance of a feature is calculated by the information gain in the building of the Decision Trees. 
We have calculate the feature importance using the training dataset with the real imbalance (8M birth certificates from 2000 and 2001). In Table~\ref{tbl:feat_bin}, we show a ranked list of the top--20 features. The top important features are: the {birth weight} and the {apgar score}. Nine features in the top--20 are highly related to the mother (we highlight those in boldface).

\section{Conclusion \& future work}
\label{sec:conclusion}

In this paper, we study the problem of infant mortality. This is a challenging \textit{binary} classification problem due to the high imbalance on the data (1:145). We show variations of training sets to tackle the problem of imbalance and we propose classification models. Our models outperforms the baselines used by researchers in Epidemiology and Medicine. We analyze the results for different mortality classes, causes of death. Also, we understand the impact of socio-economic factors in infant mortality and we build models for each race. Furthermore, we present our findings regarding the importance of the features. 
Future plans focus on extending our study to more general features from external sources, such as the spatial distribution of median household incomes, which has yet to be linked together for the prediction of infant mortality. 

\section*{Acknowledgements}
The work was funded by the H2020 LAMBDA project 734242, a Google Faculty Research Award, the Robert Wood Johnson Foundation grant 71192 and the W.K. Kellogg Foundation grant P3036220.

\bibliographystyle{ACM-Reference-Format}
\bibliography{kdd19}


\begin{thebibliography}{17}


\ifx \showCODEN    \undefined \def \showCODEN     #1{\unskip}     \fi
\ifx \showDOI      \undefined \def \showDOI       #1{#1}\fi
\ifx \showISBNx    \undefined \def \showISBNx     #1{\unskip}     \fi
\ifx \showISBNxiii \undefined \def \showISBNxiii  #1{\unskip}     \fi
\ifx \showISSN     \undefined \def \showISSN      #1{\unskip}     \fi
\ifx \showLCCN     \undefined \def \showLCCN      #1{\unskip}     \fi
\ifx \shownote     \undefined \def \shownote      #1{#1}          \fi
\ifx \showarticletitle \undefined \def \showarticletitle #1{#1}   \fi
\ifx \showURL      \undefined \def \showURL       {\relax}        \fi
\providecommand\bibfield[2]{#2}
\providecommand\bibinfo[2]{#2}
\providecommand\natexlab[1]{#1}
\providecommand\showeprint[2][]{arXiv:#2}

\bibitem[\protect\citeauthoryear{Abrevaya}{Abrevaya}{2002}]%
        {abrevaya2002effects}
\bibfield{author}{\bibinfo{person}{Jason Abrevaya}.}
  \bibinfo{year}{2002}\natexlab{}.
\newblock \showarticletitle{The effects of demographics and maternal behavior
  on the distribution of birth outcomes}.
\newblock In \bibinfo{booktitle}{{\em Economic applications of quantile
  regression}}.
\newblock


\bibitem[\protect\citeauthoryear{Acevedo-Garcia, Soobader, and
  Berkman}{Acevedo-Garcia et~al\mbox{.}}{2007}]%
        {ACEVEDOGARCIA_low_birth_weight_education}
\bibfield{author}{\bibinfo{person}{Dolores Acevedo-Garcia}, \bibinfo{person}{M.
  Soobader}, {and} \bibinfo{person}{Lisa Berkman}.}
  \bibinfo{year}{2007}\natexlab{}.
\newblock \showarticletitle{Low birthweight among U.S. Hispanic/Latino
  subgroups: The effect of maternal foreign-born status and education}.
\newblock \bibinfo{journal}{{\em Social Science \& Medicine\/}}
  (\bibinfo{year}{2007}).
\newblock


\bibitem[\protect\citeauthoryear{Acevedo-Garcia, Soobader, and
  Berkman}{Acevedo-Garcia et~al\mbox{.}}{2005}]%
        {Acevedo_Garcia2005}
\bibfield{author}{\bibinfo{person}{Dolores Acevedo-Garcia},
  \bibinfo{person}{Mah-J Soobader}, {and} \bibinfo{person}{Lisa~F. Berkman}.}
  \bibinfo{year}{2005}\natexlab{}.
\newblock \showarticletitle{The Differential Effect of Foreign-Born Status on
  Low Birth Weight by Race/Ethnicity and Education}.
\newblock \bibinfo{journal}{{\em Pediatrics\/}} (\bibinfo{year}{2005}).
\newblock


\bibitem[\protect\citeauthoryear{Almond, Chay, and Lee}{Almond
  et~al\mbox{.}}{2005}]%
        {almond2005costs}
\bibfield{author}{\bibinfo{person}{Douglas Almond}, \bibinfo{person}{Kenneth~Y
  Chay}, {and} \bibinfo{person}{David~S Lee}.} \bibinfo{year}{2005}\natexlab{}.
\newblock \showarticletitle{The costs of low birth weight}.
\newblock \bibinfo{journal}{{\em The Quarterly Journal of Economics\/}}
  (\bibinfo{year}{2005}).
\newblock


\bibitem[\protect\citeauthoryear{{Antonia Saravanou, Clemens Noelke, Nick
  Huntington, Dolores Acevedo-Garcia and Dimitrios Gunopulos}}{{Antonia
  Saravanou, Clemens Noelke, Nick Huntington, Dolores Acevedo-Garcia and
  Dimitrios Gunopulos}}{2019}]%
        {clemens2019}
\bibfield{author}{\bibinfo{person}{{Antonia Saravanou, Clemens Noelke, Nick
  Huntington, Dolores Acevedo-Garcia and Dimitrios Gunopulos}}.}
  \bibinfo{year}{2019}\natexlab{}.
\newblock \bibinfo{title}{{Predicting Infant Mortality at the Time of Birth}}.
\newblock \bibinfo{howpublished}{Population Association Annual Meeting, Austin,
  TX}.   (\bibinfo{year}{2019}).
\newblock


\bibitem[\protect\citeauthoryear{Casey, McIntire, and Leveno}{Casey
  et~al\mbox{.}}{2001}]%
        {casey2001continuing}
\bibfield{author}{\bibinfo{person}{Brian~M Casey}, \bibinfo{person}{Donald~D
  McIntire}, {and} \bibinfo{person}{Kenneth~J Leveno}.}
  \bibinfo{year}{2001}\natexlab{}.
\newblock \showarticletitle{The continuing value of the Apgar score for the
  assessment of newborn infants}.
\newblock \bibinfo{journal}{{\em New England Journal of Medicine\/}}
  (\bibinfo{year}{2001}).
\newblock


\bibitem[\protect\citeauthoryear{Chen and Guestrin}{Chen and Guestrin}{2016}]%
        {xgb}
\bibfield{author}{\bibinfo{person}{Tianqi Chen} {and} \bibinfo{person}{Carlos
  Guestrin}.} \bibinfo{year}{2016}\natexlab{}.
\newblock \showarticletitle{Xgboost: A scalable tree boosting system}. In
  \bibinfo{booktitle}{{\em Proceedings of the 22nd SIGKDD 2016}}. ACM.
\newblock


\bibitem[\protect\citeauthoryear{Doyle, Echevarria, and Frisbie}{Doyle
  et~al\mbox{.}}{2003}]%
        {doyle2003race}
\bibfield{author}{\bibinfo{person}{Jamie~Mihoko Doyle}, \bibinfo{person}{Samuel
  Echevarria}, {and} \bibinfo{person}{W~Parker Frisbie}.}
  \bibinfo{year}{2003}\natexlab{}.
\newblock \showarticletitle{Race/ ethnicity, Apgar and infant mortality}.
\newblock \bibinfo{journal}{{\em Population Research and Policy Review\/}}.
\newblock


\bibitem[\protect\citeauthoryear{Hegyi, Carbone, Anwar, Ostfeld, Hiatt, Koons,
  Pinto-Martin, and Paneth}{Hegyi et~al\mbox{.}}{1998}]%
        {hegyi1998apgar}
\bibfield{author}{\bibinfo{person}{Thomas Hegyi}, \bibinfo{person}{Tracy
  Carbone}, \bibinfo{person}{Mujahid Anwar}, \bibinfo{person}{Barbara Ostfeld},
  \bibinfo{person}{Mark Hiatt}, \bibinfo{person}{Anne Koons},
  \bibinfo{person}{Jennifer Pinto-Martin}, {and} \bibinfo{person}{Nigel
  Paneth}.} \bibinfo{year}{1998}\natexlab{}.
\newblock \showarticletitle{The Apgar score and its components in the preterm
  infant}.
\newblock \bibinfo{journal}{{\em Pediatrics\/}} (\bibinfo{year}{1998}).
\newblock


\bibitem[\protect\citeauthoryear{Hessol, Fuentes-Afflick, and Bacchetti}{Hessol
  et~al\mbox{.}}{1998}]%
        {hessol1998risk}
\bibfield{author}{\bibinfo{person}{Nancy~A Hessol}, \bibinfo{person}{Elena
  Fuentes-Afflick}, {and} \bibinfo{person}{Peter Bacchetti}.}
  \bibinfo{year}{1998}\natexlab{}.
\newblock \showarticletitle{Risk of low birth weight infants among black and
  white parents}.
\newblock \bibinfo{journal}{{\em Obstetrics \& Gynecology\/}}.
\newblock


\bibitem[\protect\citeauthoryear{Hummer, Biegler, De~Turk, Forbes, Frisbie,
  Hong, and Pullum}{Hummer et~al\mbox{.}}{1999}]%
        {hummer1999race}
\bibfield{author}{\bibinfo{person}{Robert~A Hummer}, \bibinfo{person}{Monique
  Biegler}, \bibinfo{person}{Peter~B De~Turk}, \bibinfo{person}{Douglas
  Forbes}, \bibinfo{person}{W~Parker Frisbie}, \bibinfo{person}{Ying Hong},
  {and} \bibinfo{person}{Starling~G Pullum}.} \bibinfo{year}{1999}\natexlab{}.
\newblock \showarticletitle{Race/ethnicity, nativity, and infant mortality in
  the United States}.
\newblock \bibinfo{journal}{{\em Social Forces\/}} (\bibinfo{year}{1999}).
\newblock


\bibitem[\protect\citeauthoryear{John and Langley}{John and Langley}{1995}]%
        {gnb}
\bibfield{author}{\bibinfo{person}{George~H. John} {and} \bibinfo{person}{Pat
  Langley}.} \bibinfo{year}{1995}\natexlab{}.
\newblock \showarticletitle{Estimating Continuous Distributions in Bayesian
  Classifiers}. In \bibinfo{booktitle}{{\em Proceedings of the Eleventh
  Conference on Uncertainty in Artificial Intelligence}} {\em
  (\bibinfo{series}{UAI'95})}.
\newblock


\bibitem[\protect\citeauthoryear{Kochanek, Murphy, Xu, and
  Tejada-Vera}{Kochanek et~al\mbox{.}}{2006}]%
        {cdc}
\bibfield{author}{\bibinfo{person}{Kenneth~D. Kochanek},
  \bibinfo{person}{Sherry~L. Murphy}, \bibinfo{person}{Jiaquan Xu}, {and}
  \bibinfo{person}{Betzaida Tejada-Vera}.} \bibinfo{year}{2006}\natexlab{}.
\newblock \showarticletitle{Deaths: Final Data for 2014}.
\newblock \bibinfo{journal}{{\em National Vital Statistics Reports\/}}
  (\bibinfo{year}{2006}).
\newblock


\bibitem[\protect\citeauthoryear{Osypuk and Acevedo-Garcia}{Osypuk and
  Acevedo-Garcia}{2008}]%
        {dolores_hypersegregated_areas}
\bibfield{author}{\bibinfo{person}{Theresa~L. Osypuk} {and}
  \bibinfo{person}{Dolores Acevedo-Garcia}.} \bibinfo{year}{2008}\natexlab{}.
\newblock \showarticletitle{Are Racial Disparities in Preterm Birth Larger in
  Hypersegregated Areas?}
\newblock \bibinfo{journal}{{\em American Journal of Epidemiology\/}}.
\newblock


\bibitem[\protect\citeauthoryear{Powers, Parker, and other}{Powers
  et~al\mbox{.}}{2006}]%
        {powers2006race}
\bibfield{author}{\bibinfo{person}{Daniel Powers}, \bibinfo{person}{Frisbie
  Parker}, {and} \bibinfo{person}{other}.} \bibinfo{year}{2006}\natexlab{}.
\newblock \showarticletitle{Race/Ethnic differences and age-variation in the
  effects of birth outcomes on infant mortality in the US}.
\newblock \bibinfo{journal}{{\em Demographic Research\/}}
  (\bibinfo{year}{2006}).
\newblock


\bibitem[\protect\citeauthoryear{Sch{\"o}lkopf, Williamson, Smola,
  Shawe-Taylor, and Platt}{Sch{\"o}lkopf et~al\mbox{.}}{2000}]%
        {svm}
\bibfield{author}{\bibinfo{person}{Bernhard Sch{\"o}lkopf},
  \bibinfo{person}{Robert~C Williamson}, \bibinfo{person}{Alex~J Smola},
  \bibinfo{person}{John Shawe-Taylor}, {and} \bibinfo{person}{John~C Platt}.}
  \bibinfo{year}{2000}\natexlab{}.
\newblock \showarticletitle{Support vector method for novelty detection}. In
  \bibinfo{booktitle}{{\em Advances in neural information processing systems}}.
  \bibinfo{pages}{582--588}.
\newblock


\bibitem[\protect\citeauthoryear{Wilcox and Skjaerven}{Wilcox and
  Skjaerven}{1992}]%
        {wilcox1992birth}
\bibfield{author}{\bibinfo{person}{Allen~J Wilcox} {and} \bibinfo{person}{Rolv
  Skjaerven}.} \bibinfo{year}{1992}\natexlab{}.
\newblock \showarticletitle{Birth weight and perinatal mortality: the effect of
  gestational age.}
\newblock \bibinfo{journal}{{\em American Journal of Public Health\/}}
  (\bibinfo{year}{1992}).
\newblock


\end{thebibliography}

\end{document}